\documentclass[letterpaper, 10 pt, conference]{ieeeconf} 
\usepackage{amsmath,amsfonts}
\usepackage{array}
\usepackage[caption=false,font=normalsize,labelfont=sf,textfont=sf]{subfig}
\usepackage{textcomp}
\usepackage{stfloats}
\usepackage{url}
\usepackage{verbatim}
\usepackage{graphicx}

\usepackage{amsmath, amssymb}
\usepackage{makecell}
\usepackage{multirow}
\usepackage{xcolor}         
\usepackage{amsmath}  
\usepackage{xspace}
\usepackage[ruled,linesnumbered]{algorithm2e}
\usepackage{float}

\IEEEoverridecommandlockouts

\def\BibTeX{{\rm B\kern-.05em{\sc i\kern-.025em b}\kern-.08em
    T\kern-.1667em\lower.7ex\hbox{E}\kern-.125emX}}
\usepackage{balance}
\usepackage{cite}
\usepackage[justification=centering]{caption}
\usepackage{tabularx}
\usepackage{booktabs}
\usepackage{hyperref}

\newcommand{\reflabel}{dummy} 

\newcommand{\be}{\begin{equation}}
\newcommand{\ee}{\end{equation}}
\newcommand{\eqlabel}[1]{\label{eq:\reflabel-#1}}
\renewcommand{\eqref}[2][\reflabel]{(\ref{eq:#1-#2})}

\newcommand{\seclabel}[1]{\label{sec:\reflabel-#1}}

\newcommand{\figlabel}[2][\reflabel]{\label{fig:#1-#2}}
\newcommand{\figref}[2][\reflabel]{Fig.~\ref{fig:#1-#2}}

\newcommand{\tablelabel}[2][\reflabel]{\label{table:#1-#2}}
\newcommand{\tableref}[2][\reflabel]{Table~\ref{table:#1-#2}}

\newcommand{\ie}{i.e.\xspace}

\newcommand{\etal}{et al.\xspace}

\newcommand{\uncert}{\sigma}

\usepackage{hyperref}

\let\oldtwocolumn\twocolumn
\renewcommand\twocolumn[1][]{%
    \oldtwocolumn[{#1}{
    \begin{center}
    \includegraphics[trim=0 1 0 5, clip, width=1.0\textwidth]{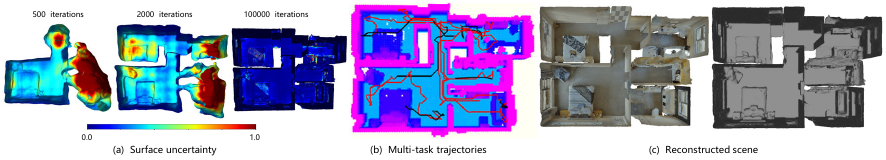}
      \captionof{figure}{(a) Surface uncertainty from reconstruction iterations; (b) Planned multi-task trajectories (The black line represents exploration tasks, and the red line represents merged tasks); (c) Reconstructed mesh with texture and without texture.}
    \label{teaser}
    \end{center}
    }
    ]
}

\begin{document}

\title{Autonomous Implicit Indoor Scene Reconstruction with Frontier Exploration}
\author{Jing Zeng$^{1}$$^{\dagger}$, Yanxu Li$^{1}$$^{\dagger}$, Jiahao Sun$^{1}$$^{\dagger}$, Qi Ye$^{1}$$^{*}$, Yunlong Ran$^{1}$, Jiming Chen$^{1}$  
\thanks{$\dagger$ $\textbf{Equal contribution}$}
\thanks{$^{1}$ College of Control Science and Engineering, Zhejiang University, Hangzhou, 310027, China.}
\thanks{$^{*}$ Qi Ye (Corresponding author, qi.ye@zju.edu.cn) is with the College of Control Science and Engineering, the State Key Laboratory of Industrial Control Technology, Zhejiang University, and the Key Key Lab of CS\&AUS of Zhejiang Province.}
\thanks{This work was supported in part by the National Natural Science Foundation of China (Grant Number: 62088101, 62233013, 62103372 ).}
}
\maketitle


\begin{abstract}

Implicit neural representations have demonstrated significant promise for 3D scene reconstruction. Recent works have extended their applications to autonomous implicit reconstruction through the Next Best View (NBV) based method. However, the NBV method cannot guarantee complete scene coverage and often necessitates extensive viewpoint sampling, particularly in complex scenes. 
In the paper, we propose to 1) incorporate frontier-based exploration tasks for global coverage with implicit surface uncertainty-based reconstruction tasks to achieve high-quality reconstruction. and 2) introduce a method to achieve implicit surface uncertainty using color uncertainty, which reduces the time needed for view selection. Further with these two tasks, we propose an adaptive strategy for switching modes in view path planning, to reduce time and maintain superior reconstruction quality. Our method exhibits the highest reconstruction quality among all planning methods and superior planning efficiency in methods involving reconstruction tasks. We deploy our method on a UAV and the results show that our method can plan multi-task views and reconstruct a scene with high quality. 


\end{abstract}

\section{Introduction}

Reconstructing and mapping indoor environments is crucial for various applications, including scene visualization, robot navigation, and 3D content creation for augmented and virtual reality~\cite{guo2022asynchronous,xiang2023nisb,zhu2022nice,shu19infocom}. In recent years, the emergence of compact and agile aerial vehicles, such as UAVs, has sparked a growing interest in scene reconstruction using close-range aerial imagery. 



Implicit neural representations for 3D objects have demonstrated considerable promise in scene reconstruction, scene editing, and robotics and autonomous systems~\cite{yu2022monosdf,wang2023seal,zeng2023efficient}. By leveraging the expressive power of implicit representations, autonomous systems can reconstruct 3D environments online and plan optimal view paths for data acquisition~\cite{pan2022activenerf,ran2023neurar,zeng2023efficient,shu16tiegloc}. Ran~\etal~\cite{ran2023neurar} first proposes a novel view quality criterion based on color uncertainty that is learned online for view selection. To reduce the computational complexity in querying viewpoint-specific information gain, Zeng~\etal~\cite{zeng2023efficient} introduces an implicit function approximator for the information gain field. However, these methods employ a greedy planning strategy based on the NBV, which is prone to result in local optimum when reconstructing large scenes, resulting in incomplete scene coverage. 

Scene exploration methods\cite{zhou2021fuel,guo2022asynchronous,dong2019multi} where agents are required to reach maximum coverage of target scenes find that incorporating global information helps to escape from the local minima. Inspired by these methods, we propose to leverage the global information in the autonomous implicit reconstruction pipeline. 
Particularly, we incorporate frontier-based exploration tasks with autonomous implicit reconstruction tasks, combining the benefits of both approaches to enhance the overall efficiency and efficacy of reconstruction.


However, the incorporation faces two main challenges. 
First, as the scene size expands, the expenses for sampling a wider range of viewpoints and calculating information gained for each viewpoint increase with the scene volume.
Second, the coordination of exploration tasks and reconstruction tasks requires a trade-off between efficiency and efficacy. Generating reconstruction tasks based on implicit representation usually takes orders of time complexity than generating exploration tasks. Including reconstruction tasks in every iteration of planning will lead to an increase in planning time while
merely switching between exploration tasks and reconstruction tasks in each iteration can result in poor surface quality or repeatedly getting trapped in local optimum during the scanning process.



For the first challenge, we observe that the reconstruction uncertainty of the space converges to very low values in a few iterations and the high uncertainty (or poor reconstruction quality) mainly happens near surface areas. Therefore, instead of sampling the viewpoints in all the space and calculating the Next-Best-View information gain by integrating the color uncertainty for 3D points in each viewpoint frustum, we propose to evaluate the surface quality, only sample viewpoints covering surfaces of low quality and sum the surface uncertainty in a viewpoint, similar to surface inspection methods~\cite{bircher2018receding,guo2022asynchronous,kompis2021informed}, which reduces the complexity measuring based on 3D volume to 2D surface. 



For the second challenge, we propose an adaptive mode-switching approach for view path planning based on the number of frontiers within the current neighborhood. The design comes from the intuition that when an agent enters a new area, exploration tasks dominate; when a rough overview of the area is gathered, reconstruction tasks are included to focus on finer details. To avoid getting trapped in local regions, exploration tasks are planned together with the reconstruction tasks in the latter state.  With this model-switching approach, our method achieves low time cost and high-quality reconstruction. 

To summarize, our contributions are:
\begin{itemize}
    \item We incorporate frontier-based exploration tasks for efficient global coverage with implicit surface uncertainty based reconstruction tasks for high-quality reconstruction.
    \item We propose a novel information gain calculation and corresponding viewpoint sampling strategy that evaluates the uncertainty of implicit surface and samples in the space covering the high uncertainty surface areas. 
    \item We propose an adaptive mode-switching approach for view path planning to coordinate exploration and reconstruction tasks.
\end{itemize}

\section{Related work}

Implicit representations, exemplified by NeRF, have emerged as a powerful approach for 3D scene reconstruction. There are numerous variations of NeRF, with some focusing on appearance, such as TensoRF~\cite{chen2022tensorf} and Instant-NGP~\cite{muller2022instant}, some focusing on geometry, such as Neus~\cite{wang2021neus}, VolSDF~\cite{yariv2021volume}, and MonoSDF~\cite{yu2022monosdf}.

View path planning is a crucial aspect of autonomous reconstruction, aiming to determine an optimal sequence of viewpoints to efficiently reconstruct a 3D scene~\cite{kompis2021informed,schmid2020efficient,chen2022real}. Next best view (NBV) and frontier-based methods have emerged as popular approaches for autonomous reconstruction. NBV-based methods employ a greedy strategy, selecting the viewpoint with the highest expected gain, which is expected to provide the most valuable information for the reconstruction process~\cite{pan2021global,zeng2023efficient}. Despite their effectiveness, the challenge of converging to local optimum within a confined region is a common issue encountered by NBV-based methods~\cite{wang2019autonomous,almadhoun2019guided}. This issue can be attributed to the constraints imposed by limited sampling range and resolution, which restrict the exploration of the full solution space. Frontier-based methods, on the other hand, prioritize the exploration of unknown regions in the scene through the identification and selection of frontier points, which denote the boundaries between known and unknown areas~\cite{zhou2021fuel,kompis2021informed,dong2019multi}. However, it is worth highlighting that the mere coverage of these frontiers does not guarantee improved reconstruction quality. This characteristic induces frontier-based methods more suitable for autonomous exploration rather than reconstruction.

Some studies~\cite{song2021view,guo2022asynchronous} propose a combination of NBV-based and frontier-based methods. Song et al.~\cite{song2021view} select the nearest frontier task as NBV and analyze the quality of reconstructed surfaces. The resulting computed path achieves full coverage of low-quality surfaces. However, solely choosing the nearest frontier task tends to reduce exploration speed. Guo et al.~\cite{guo2022asynchronous} introduce an asynchronous collaborative autonomous scanning approach featuring mode switching, effectively utilizing multiple robots to explore, and reconstruct unknown scene environments. However, in autonomous implicit reconstruction, there is currently no method that combines these two tasks.

Autonomous implicit reconstruction has seen recent advancements~\cite{ran2023neurar,zeng2023efficient,yan2023active}. Ran et al.~\cite{ran2023neurar} propose neural uncertainty as a view quality criterion and the first autonomous implicit reconstruction system. Zeng et al.~\cite{zeng2023efficient} introduce a view information gain field to reduce the time for view selection. Yan et al.~\cite{yan2023active} provide a new perspective of active mapping from the optimization dynamics of map parameters.

\section{Method}

\seclabel{problem}

\subsection{Problem Statement and System Overview} 
The primary focus of this study is to address the challenge of exploring unknown and spatially bounded 3D environments while reconstructing high-quality 3D models using a mobile robot. In previous greedy-based NBV methods\cite{ran2023neurar,zeng2023efficient}, the whole reconstruction process consists of multiple iterations, and in each iteration, viewpoints are sampled and evaluated to choose the best view for the next iteration. Different from these methods, we incorporate a frontier-based exploration strategy into the reconstruction task: the whole process consists of a series of exploration and reconstruction tasks and the two subtasks interweave in or switch between reconstruction iterations. 

\begin{figure*}
\centering
\subfloat[]{
		\includegraphics[width=0.2\linewidth, height=0.16\linewidth]{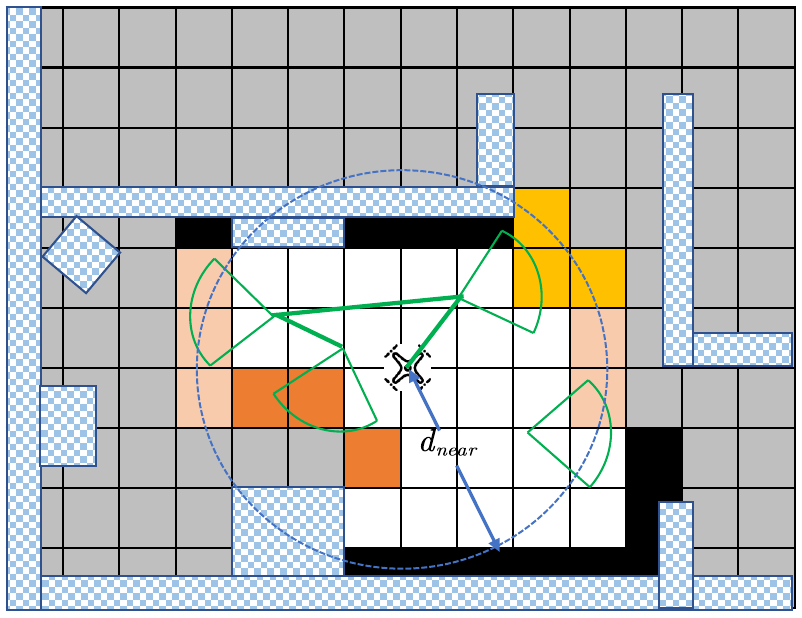}}
\subfloat[]{
		\includegraphics[width=0.2\linewidth, height=0.16\linewidth]{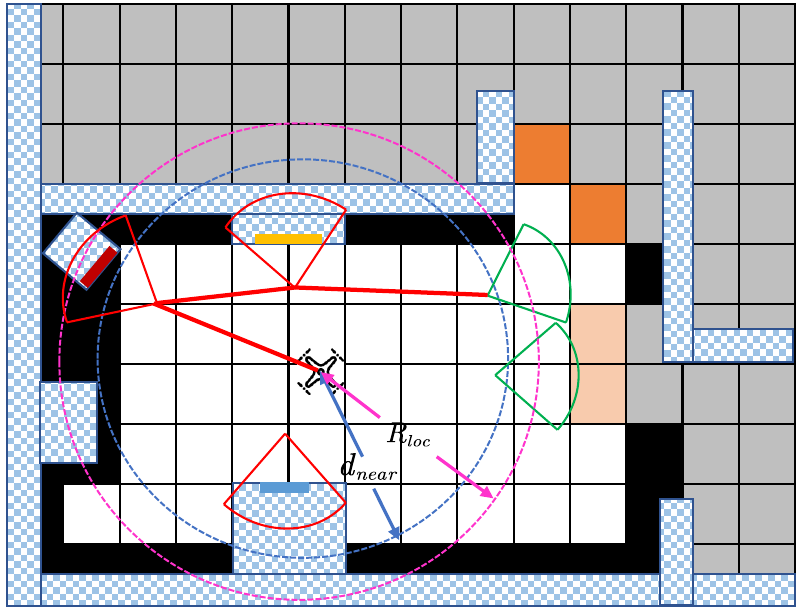}}
\subfloat[]{
		\includegraphics[width=0.25\linewidth, height=0.16\linewidth]{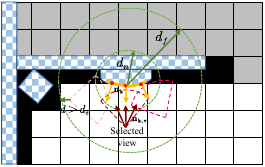}}
\subfloat{
		\includegraphics[width=0.3\linewidth, height=0.16\linewidth]{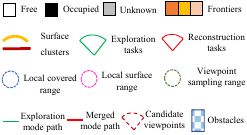}}
\caption{Overview of our method. Views and paths are planned in exploration mode (a) and merged mode (b). Candidate view sampling for each surface cluster is depicted in (c).}
\label{method_details}
\vspace{-7mm}
\end{figure*}

Under the framework of the multi-task strategy, our pipeline is composed of three components, as illustrated in \figref{pipeline}. The method overview is shown in \ref{method_details}. \textbf{The Mobile Robot module} captures images at specified viewpoints and utilizes Droid-SLAM~\cite{teed2021droid} to estimate its localization. During the simulation, Unity Engine renders images at given viewpoints similar to \cite{zeng2023efficient}.  
\textbf{The Mapping module} maintains two representations. For exploration tasks, a volumetric representation ( occupancy grid map) is adopted and for reconstruction tasks, an implicit neural representation (MonoSDF\cite{yu2022monosdf}) is adopted where monocular depth and normal are derived via Omnidata\cite{eftekhar2021omnidata}. 
\textbf{In the View Path Planning module} the number of frontiers extracted from the current volumetric map determines the optimal mode: exploration or the combination of both exploration and reconstruction at each reconstruction iteration. Leveraging information extracted from the volumetric map and the partial reconstruction scene, exploration and reconstruction tasks are generated accordingly. An informative path, incorporating multiple tasks, is planned using the Asymmetric Traveling Salesman Problem (ATSP) planner. The images captured along this path are then fed into the reconstruction system until the entire autonomous reconstruction process is completed.


\begin{figure}[htbp]
\vspace{-2mm}
    \centering
\includegraphics[width=1.0\linewidth]{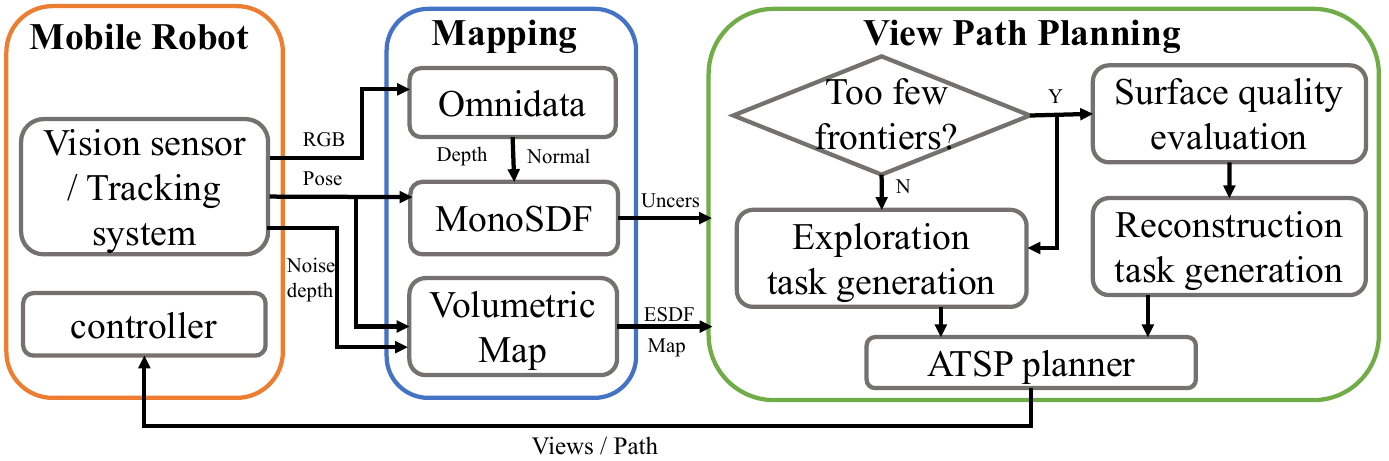}
\caption{The pipeline of our proposed method.} 
\figlabel{pipeline}
\vspace{-6mm}
\end{figure}

 
\subsection{Implicit scene reconstruction with surface uncertainty}

In this research, we employ MonoSDF~\cite{yu2022monosdf} with multi-grid features and monocular priors, as the foundational framework for our investigation. In alignment with the NeurAR~\cite{ran2023neurar} approach, we depict the rendered colors through Gaussian distributions, wherein the variance serves as a quantifier of viewpoint-related uncertainty.

Benefiting from the accurate surface characteristics of MonoSDF, we can quickly extract the surface and analyze the surface uncertainty of the reconstruction. This procedure can be decomposed into two steps. Firstly, we partition the space into $N_o^3$ grids, where $N_o=256$, then we employ the implicit grids network $F$ to compute the sdf $s_x$, gradients $g_x$, and features $\hat{z}_x$ for each grid $x$ as follows
\be
s_x, g_x, \hat{z}_x = F(x)
\eqlabel{implicit_network}
\end{equation}
Following that, by utilizing the zero-level Marching Cubes algorithm~\cite{lorensen1998marching}, we can extract the scene surface $S$ and all the vertices from $S$. Secondly, due to our emphasis on surface uncertainty evaluation, we simplify the volume rendering process by disregarding ray directions towards surface points and can obtain surface point color $c$ and uncertainty $\sigma$ of each surface point $x_s$ from rendering network $C$ as follows
\be
c_{x_s}, \sigma_{x_s} = C(x_s,g_{x_s},\hat{z}_{x_s})
\eqlabel{render_network}
\end{equation}

\subsection{Multi-task generation}
Within the framework of our approach, the scanning tasks can be divided into two tasks: exploration tasks for rapid coverage and reconstruction tasks for high-quality reconstruction. 

\subsubsection{Exploration task generation}
The exploration tasks $T^{exp} = \{T_1^{exp}, T_2^{exp}, ..., T_{N_{exp}}^{exp}\}$ are designed to cover more unknown regions, with $N_{exp}$ denoting the total number of tasks. Each exploration task can be denoted as $T_i^{exp}=(p_i^{exp},\theta_{i}^{exp},\phi_{i}^{exp})$, where $p_i$ is the 3D position, $\theta_{i}^{exp} \in (-\frac{\pi}{2}, \frac{\pi}{2})$ represents the pitch, and $\phi_{i}^{exp} \in [-\pi, \pi)$ represents the yaw of the robot.

To explore unknown regions more effectively, we select viewpoints that can provide superior coverage of the frontiers as the exploration tasks. Specifically, we firstly update incremental frontiers and Euclidean Signed Distance Field (ESDF) map $E$~\cite{han2019fiesta} from the maintained volumetric map $V=V_o \cup V_e \cup V_u$ similar to Fuel~\cite{zhou2021fuel}, where $V_o, V_e, V_u$ represent occupied, empty and unknown voxels. The ESDF map is responsible for filtering out viewpoints that are too close to obstacles. Secondly, in contrast to Fuel, to achieve more comprehensive coverage, we extract 3D principal components using Principal Component Analysis (PCA) instead of 2D and split large frontier clusters along the directions of the principal components. 

Finally, for each frontier cluster $F_i$, we generate a series of candidate viewpoints $VP_i^{exp}=\{v_{i,1}^{exp},v_{i,2}^{exp},...,v_{i,n}^{exp}\}$, where $v_{i,j}^{exp}=(p_{i,j}^{exp},\theta_{i,j}^{exp},\phi_{i,j}^{exp})$. $VP_i^{exp}$ are uniformly sampled on the spherical shell centered at the cluster, within the space $Ve$, and oriented towards the center. The radius of the spherical shell ranges from $d_r$ to $d_f$ while ensuring that the distance to the nearest obstacle from $E$ is greater than $d_s=0.3m$. Similar to~\cite{witting2018history}, We select the candidate viewpoint $v_i^{exp}$ with the largest number of visible cluster cells as the exploration task $T_i^{exp}=v_i^{exp}$ for the frontier cluster.

\subsubsection{Reconstruction task generation}
The reconstruction tasks $T^{rec} = \{T_1^{rec}, T_2^{rec}, ..., T_{N_{rec}}^{rec}\}$ are responsible for refining surface with lower quality, with $N_{rec}$ denoting the total number of tasks. Each exploration task can be denoted as $T_i^{rec}=(p_i^{rec},\theta_{i}^{rec},\phi_{i}^{rec})$ with the same range as exploration tasks. Algorithm \ref{alg:rec} describes the generation process of reconstruction tasks.

        
\vspace{-3mm}
\begin{algorithm}\footnotesize
\caption{Reconstruction task generation}
\label{alg:rec}
\SetKwInOut{Input}{Surface clustering hyperparameters} 
\Input{local surface \par radius $R_{loc}$, clustering radius $R_{clu}$, num $N_{rec}$}
\SetKwInOut{Input}{View sampling hyperparameters}
\Input{sample range \par$d_r$ $d_f$, nearest distance $d_s$, visible threshold $N_{min}$}
\KwIn{Implict surface model $S$, volumetric map $V$, ESDF map $E$, current position $p_0$}
\KwOut{Updated reconstruction tasks $T^{rec}$}
$S_d \leftarrow SurfaceDownsampling(S, N_{down})$ \;
$S_v \leftarrow  LocalSurfaceSampling(S_d, R_{loc})$ \;
$U \leftarrow LocalSurfaceClustering(S_v, R_{clu}, N_{rec})$ \;
\For{$U_i \in U$}{
     \tcp{Iterate over surface clusters}
     $VP_i^{rec} \leftarrow ViewSampling(V,U_i,d_r, d_f, d_s,N_{min}, p_0)$ \;
     $v_i^{rec} \leftarrow ViewofMaxGain(VP_i^{rec})$ \;
     $T_i^{rec} \leftarrow v_i^{rec}$
}
$T^{rec} \leftarrow \{T_1^{rec}, T_2^{rec}, ..., T_{N_{rec}}^{rec}\}$
\end{algorithm}
\vspace{-3mm}

\noindent\textbf{Surface Downsampling}
To enhance the efficiency of detecting low-quality surfaces, we perform $N_{down}=5$ times by downsampling the surface $S$, leading to the generation of a new set of surface elements represented as $S_d = \{s_1, s_2, ..., s_{N_d}\}$, consisting of $N_d$ elements. Each surface element is defined as  $s_j = \{x_j,\mathbf{n}_j,\sigma_j\}$, where $x_j$ is the j-th vertex on the surface. With Marching Cubes and rendering network in \eqref{render_network}, we can obtain uncertainty $\sigma_j$, and normal vector $\mathbf{n_j}$ for each surface vertex $x_j$.

\noindent\textbf{Local Surface Sampling and Clustering}
Assuming the current position of the robot is $p_0$, we select all surface elements $S_v \in S_d$ within a small radius $R_{loc}$ from $p_0$ to ensure that the generated reconstruction tasks remain within a local scope. Considering the substantial number of surface elements, we can group these elements $S_v$ into $N_{rec}$ surface clusters $U = \{U_1, U_2, ..., U_{N_{rec}}\}$. This allows us to simplify the process of generating viewpoints from different clusters, reducing redundant computations. We adopt an uncertainty-guided iterative clustering approach. Initially, from the surface elements $S_v$, we select the element with the highest uncertainty as the clustering center. We then include all points within a radius $R_{clu}$ of this center in one cluster. Subsequently, we repeat the above process on the remaining surface elements until $N_{rec}$ clusters are generated. 

\noindent\textbf{View Sampling}
For each surface cluster $U_i$, we compute its center and generate a series of candidate viewpoints $VP_i^{rec}=\{v_{i,1}^{rec},v_{i,2}^{rec},...,v_{i,m}^{rec}\}$, where $v_{i,j}^{rec}=(p_{i,j}^{rec},\theta_{i,j}^{rec},\phi_{i,j}^{rec})$. These viewpoints in $VP_i^{rec}$ are uniformly sampled on the spherical shell centered at the cluster within the empty space $Ve$, and they are oriented towards the cluster center. We also ensure that the count of visible cluster cells from these viewpoints exceeds the threshold $N_{min}$. The radius of the spherical shell ranges from $d_r$ to $d_f$ while ensuring that the distance to the nearest obstacle from $E$ is greater than $d_s$.

\noindent\textbf{Surface Uncertainty Based Information Gain} To select reconstruction tasks from these candidate viewpoints, we define the viewpoint information gain as 
\be
g(v) = \sum_{k=1}^{N_{vis}} |\mathbf{n}_{k,v} \cdot \mathbf{n}_k| \sigma_k
\ee
where  $N_{vis}$ represents the number of visible surface elements, $\sigma_k$ represents the uncertainty of each visible surface element, and $\mathbf{n}_{k,v}$ is the vector from the viewpoint $v$ to the surface element $s_k$. Notice that the proposed surface uncertainty based information gain differs from the viewpoint information gain in \cite{ran2023neurar,zeng2023efficient}
\be
g(v) = \frac 1 R \sum_{r=1}^R\sum_{i=1}^{N}\omega_{ri} \sigma_{ri}^2,
\eqlabel{view_uncer}
\ee
where  $\uncert_{ri}^2$ is the uncertainty of the color for a 3D point sampled on a ray $r$ tracing through an image pixel of a viewpoint. $R$ is the number of sampled rays, $N$ is the number of sampled points on each ray, and $w_{ri}$ is the weight. 
The viewpoint information gain integrates all the 3D points sampled in a viewpoint frustum while our information gain only integrates points on 2D surfaces. The surface uncertainty and the surface geometry also help to reduce the viewpoint sampling space.

We then choose the viewpoint $v_i^{rec}$ with the highest information gain as the reconstruction task $T_i^{rec}=v_i^{rec}$ for this surface cluster.

\subsection{Adaptive view path planning}

As aforementioned, the allocation of exploration and reconstruction tasks faces the challenges of trade-off between efficiency of efficacy. To address these challenges, we design a mode-switching approach for view path planning to achieve both planning efficiency and high reconstruction quality: when many frontiers are to be explored,  exploration tasks dominate; otherwise, exploration and reconstruction tasks are planned together to get finer details and also avoid getting trapped in local regions.

Specifically, the switch condition depends on 1) the number $N_{near}$ of frontiers in the neighborhood of the current agent position $p_0$,  the neighborhood being a sphere located within a distance of $d_{near}$ from $p_0$; 2) the ratio $\beta_{near}$ of the frontiers in this local region to all the frontiers remained. 
When $N_{near} > 3$ and $\beta_{near} > 0.2$, the robot performs exploration tasks, i.e. $T = T^{exp}$; otherwise, the robot will simultaneously perform both exploration and reconstruction tasks, i.e. $T = T^{exp} \cup T^{rec}$. Similar to~\cite{zhou2021fuel}, our problem becomes an ATSP Problem:
\be
\vspace{-0.8mm}
T^* = \ arg\ \min \sum_{T_k \in T} d(T_k,T_{k+1})
\eqlabel{ATSP}
\vspace{-0.8mm}
\end{equation}

where $ d(T_k,T_{k+1})$ is the length of A* path search from task $T_k$ to $T_{k+1}$. $T^*$ represents the planned sequence of tasks. 

As the map status updates with tasks executed, we choose to perform planned tasks in  $T^*$ whose path length from the agent's current position is smaller than a threshold $L_{exec}$ to form the final execution sequence $T^*_e = \{T_1, T_2, ..., T_{N_t}\}$. 
Finally, we uniformly sample viewpoints along the sequence $T^*_e$ at a path resolution of $l_{res}=0.2m$ and capture images as inputs for the mapping process. It is worth noting that when all frontiers are fully covered and exploration tasks are completed, only reconstruction tasks remain. At this stage, we increase local sampling radius $\alpha R_{loc}$, clustering radius $\alpha R_{clu}$, clustering num $\alpha N_{rec}$, preceding path length $\alpha L_{exec}$ and path resolution $\alpha l_{res}$ until a certain number of images are collected, where $\alpha$ = 3 is the amplification coefficient.

\section{Results}

\begin{table*}[!ht]
    \vspace{2mm}
    \centering
    \caption{Evaluations of the effectiveness and efficiency of view path for implicit surface representation.}
    \resizebox{\textwidth}{25 mm}{
    \normalsize
    \setlength{\tabcolsep}{1mm}{
    \begin{tabular}{c|cccc|cccc|cccc|cccc}
    \toprule
         \multicolumn{1}{c}{}& \multicolumn{4}{c}{Method} & \multicolumn{4}{c}{Bhxhp} & \multicolumn{4}{c}{Rosser} & \multicolumn{4}{c}{Convoy}  \\ 
        Variant & Pitch & $T_{exp}$ & $T_{rec}$ & Switch & PSNR↑ & Acc↓ & Comp↓ & Recall↑ & PSNR↑ & Acc↓ & Comp↓ & Recall↑  & PSNR↑ & Acc↓ & Comp↓ & Recall↑ \\ 
        \midrule
        
        V1(Fuel\textsuperscript{\cite{zhou2021fuel}}) & & \checkmark  &  &   & 27.26 & 2.42 & 2.77 & 0.87  & 16.85 & 4.01 & 3.83 & 0.77 & 19.03 & 4.07 & 2.56 &  0.89   \\ 
        V2 & \checkmark& \checkmark &  &  & 28.28 & 2.25 & 2.67 & 0.88 & 20.33 & 2.93 & 2.48 & 0.91 & 20.03 & 3.74 & 2.21 & 0.93 \\
        V3 & \checkmark&  & \checkmark &  & 16.50 & 35.09 & 49.10 & 0.23      & 14.87 & 9.44 & 14.96 & 0.54 & 16.03 & 8.26 & 19.59 & 0.56 \\
        V4 & \checkmark & \checkmark & \checkmark &   & 26.66  & 17.87 & 13.32 & 0.73 & 18.87 & 4.87 & 4.05 & 0.79 & 20.59 & 5.07 & 3.98 & 0.90  \\
        V5(Ours full) & \checkmark & \checkmark & \checkmark & \checkmark  & \textbf{30.45} & \textbf{1.90} & \textbf{2.18} & \textbf{0.92} & \textbf{22.04} & \textbf{2.76} & \textbf{2.27} & \textbf{0.94} & 
        \textbf{21.24} & \textbf{3.24} & \textbf{1.96} & \textbf{0.96}\\  
     
        \midrule
         Variant & Pitch & $T_{exp}$ & $T_{rec}$ & Switch & $T_{task}$  & $T_{SP}$ & $T_{GP}$ & P.L.  & $T_{task}$  & $T_{SP}$ & $T_{GP}$ & P.L. & $T_{task}$  & $T_{SP}$ & $T_{GP}$ & P.L.\\ \midrule
        V1(Fuel\textsuperscript{\cite{zhou2021fuel}}) & & \checkmark  &  &  &  0.004  & 0.025 & 0.78 & 57.08 & 0.005 & 0.036 & 0.46 & 27.54 & 0.005 & 0.035 & 0.26 & 15.72\\
        V2 & \checkmark& \checkmark &  &  &  0.012  & 0.040 & 1.71 & 55.96 & 0.021 & 0.050 & 0.83 & 27.03 & 0.022 & 0.049 & 0.33 & 13.75\\
        V3 & \checkmark&  & \checkmark & & 3.12  & 3.13 & 89.66 & 104.30 & 3.077 & 3.091 & 48.02 & 47.62 & 3.13 & 3.15 & 29.88 & 31.29\\
        V4 & \checkmark & \checkmark & \checkmark &  &  3.12  & 3.18 & 46.78 & 109.09  &  3.13  & 3.17 & 28.80 & 53.42 & 3.05 & 3.09 & 13.88 & 34.55 \\
        V5(Ours full) & \checkmark & \checkmark & \checkmark & \checkmark & 0.01 / 3.33  & 0.03 / 3.37 & 40.02 & 108.23 & 0.01 / 3.06  & 0.04 / 3.11 & 26.19 & 55.56 &
        0.01 / 2.93 & 0.04 / 2.96 & 8.41 & 40.56\\
     
         \midrule
    \end{tabular} 
    } 
    }
    \tablelabel{table_effect_effici}
     \vspace{-2mm}
\end{table*}

\begin{table*}[ht]
    \vspace{-2mm}
    \centering
    \caption{Evaluations of the effectiveness and efficiency with existing planning methods.}
    \resizebox{\textwidth}{11mm}{
    \normalsize
    \setlength{\tabcolsep}{1.0mm}{
    \begin{tabular}{c|cccccc|cccccc|cccccc}
    \toprule
        \multicolumn{1}{c}{} & \multicolumn{6}{c}{Bhxhp} & \multicolumn{6}{c}{Rosser}&\multicolumn{6}{c}{Convoy}   \\ 
        Method  & PSNR↑ & Acc↓ & Comp↓ & Recall↑ & $T_{GP}$ & P.L. & PSNR↑ & Acc↓ & Comp↓ & Recall↑ & $T_{GP}$ & P.L. & PSNR↑ & Acc↓ & Comp↓ & Recall↑ & $T_{GP}$ & P.L. \\ 
        \midrule
        EVPP\textsuperscript{\cite{zeng2023efficient}} & 21.83 & 27.56 & 34.26 & 0.53 & 2138.93 & \textbf{47.75} & 19.74 & 4.78 & 3.25 & 0.86 & 973.81 & \textbf{26.58} & 18.83 & 7.21 & 4.52 & 0.84 & 933.89 & \textbf{15.06}\\ 
        
         VPP\textsuperscript{\cite{song2021view}} & 24.01 & 24.36 & 31.99 & 0.54 & 59.62 & 109.18 & 20.00 & 4.77 & 3.95 & 0.86 & 29.37 & 57.22 & 20.60 & 7.20 & 4.45 & 0.87 & 18.12 & 37.82\\ 
         
         Our & \textbf{30.45}  & \textbf{1.90} & \textbf{2.18} & \textbf{0.92} & \textbf{40.02} & 108.23 & \textbf{22.04} & \textbf{2.76} & \textbf{2.27} & \textbf{0.94} & \textbf{26.19} & \textbf55.56 & \textbf{21.24} & \textbf{3.24} & \textbf{1.96} & \textbf{0.96} & \textbf{8.41} & 40.56

          \\
        \midrule
    \end{tabular}
    }
    }
    \tablelabel{compare_with_existing_methods}
    \vspace{-4mm}
\end{table*}


\subsection{Implementation details}
\seclabel{impd}

\subsubsection{Data}
The experiments are conducted on three virtual scenes: $Bhxhp$ ($65m^2$) from HM3D~\cite{ramakrishnan2021hm3d}, $Rosser$ ($40m^2$) and $Convoy$ ($20m^2$) from Gibson~\cite{xiazamirhe2018gibsonenv}, which are reconstructed and modified from real scenes. In order to minimize unproductive exploration beyond the designated scene boundaries, we have sealed all windows within the scenes. We maintain the same depth noise and field of view (FOV) parameters, as specified in prior works such as~\cite{ran2023neurar,zeng2023efficient}.

\subsubsection{Implementation}

Our method runs on two RTX3090 GPUs. The implicit surface reconstruction is on one GPU, Omnidata and view path planning is on the other one, where view path planning runs in the ROS environment. We set the maximum planned views to be 250 views for $Bhxhp$, 130 views for $Rosser$, and 80 views for $Convoy$. All scene-dependent parameters are listed in \tableref{table_param}.

\begin{table}[htbp]
    \vspace{-2mm}
    \centering
    \caption{Scene-dependent parameters.}
    \tablelabel{table_param} 
    \resizebox{75mm}{9mm}{
    \normalsize
    \setlength{\tabcolsep}{1.0mm}{
    \begin{tabular}{ccccccccc}
    \toprule
        Scene & $d_r$ & $d_f$ & $R_{loc}$  & $R_{clu}$ & $N_{rec}$ & $N_{min}$ & $d_{near}$ & $L_{exec}$  \\
         \midrule
        Bhxhp & 1.0 & 2.0 & 2.5  & 1.3 & 15 & 30 & 2.0  & 6.0  \\
        Rosser & 0.8 & 1.8 & 2.0  & 1.1 & 10 & 20 & 1.5  & 5.0   \\
        Convoy & 0.5 & 1.5 & 1.5  & 1.0 & 10 & 15 & 1.0  & 4.0  \\
        \midrule
    \end{tabular}
    }
    }
    \vspace{-3mm}
\end{table}

\subsubsection{Metric} Similar to~\cite{zeng2023efficient}, we evaluate our method from two aspects including effectiveness and efficiency. The quality of reconstructed scenes is measured in two parts: the quality of the rendered images and the quality of the geometry of the reconstructed surface. We adopt metrics from MonoSDF~\cite{yu2022monosdf}: Accuracy (cm), Completion (cm), Recall (the percentage of points in the reconstructed mesh with Completion under 5 cm. For the geometry metrics, about 300k points are sampled from the surfaces.

For efficiency, we evaluate the path length (meter) and the planning time (second). The total path length is $P.L.$ and the time is $T_{GP}$. For the time of our view path planning for each step of the reconstruction process, we break it into several parts for more detailed comparison: 1) the task generation time $T_{task}$, 2) the time $T_{atsp}$ for the ATSP based view path planning, 3) the mode switching time $T_{switch}$ during the planning. We also report the total view planning time for each step $T_{SP}$ \ie $T_{SP}=T_{task}+T_{atsp}+T_{switch}$. 
$T_{atsp}$ for all the variants in ~\tableref{table_effect_effici} is about 0.02s to 0.06s. $T_{switch}$ is about 0.003s.

\subsection{Efficacy of the Method} 
Similar to~\cite{zeng2023efficient}, the efficacy of the method is evaluated regarding both
the effectiveness and efficiency of our contributions. We design variants of our method based on implicit surface representation. The variants are V1 (Fuel~\cite{zhou2021fuel}), V2 (Fuel with pitch), V3 (only reconstruction tasks), and V4 (Merging exploration and reconstruction tasks without mode switching).
Our method is best for effectiveness and better than methods with reconstruction tasks for efficiency. 

\subsubsection{Uncertainty of implicit surface}
Fig. \ref{teaser} illustrates the evolution of surface uncertainty during the training process. It can be observed that surface uncertainty gradually decreases during the training process. The high surface uncertainty in regions with poor geometric structures allows us to allocate viewpoints for scanning these low-quality surfaces. This reduces the time required for viewpoint sampling compared to NBV methods such as~\cite{zeng2023efficient}.

\subsubsection{Combination of exploration and reconstruction tasks}
We make V1 our baseline, which uses incremental frontiers to explore unknown environments. We make V2 with a pitch angle to verify its efficacy. To verify the efficacy of different tasks, we make V2 (exploration tasks only) and V3 (reconstruction tasks only) as our baselines.

The metrics of V2 in \tableref{table_effect_effici} demonstrate pitch angle can improve reconstruction quality than \textbf{Fuel}~\cite{zhou2021fuel} and introduce a slight increase in planning time. Focusing solely on exploration tasks (V2) results in rapid scene coverage, but it fails to perform further detailed scans of complex details, thus reducing the reconstruction quality. When only reconstruction tasks are considered (V3), the reconstruction quality becomes very poor because it falls into the local optimum and cannot cover the entire scene. It is typical for the incorporation of reconstruction tasks to lead to an increase in planning time, as it requires surface uncertainty extraction, similar to other surface inspection methods such as~\cite{guo2022asynchronous}.

\subsubsection{Adaptive view path planning}
To validate the effectiveness of mode switching, we establish V4 as our baseline. However, the merger of these two tasks significantly slows down the pace of scene exploration, and even leads to local optimum, especially when dealing with complex reconstruction details. The introduction of mode switching (Ours) ensures the speed of exploration and scanning of details, without falling into local optimum associated with scanning low-quality surfaces.

\subsection{Comparison with existing planning methods}
\seclabel{cwepm}
We select two recent works \textbf{EVPP}~\cite{zeng2023efficient} based on view information gain filed and \textbf{VPP}~\cite{song2021view}, which uses the nearest exploration task as the NBV and plans a path to the NBV using surface inspection methods. The metrics in \tableref{compare_with_existing_methods} show our method outperforms them in the reconstructed quality and the planning efficiency. This is because \textbf{EVPP} fails to avoid local minima and cannot explore the entire scene, \textbf{VPP} only relies on the nearest exploration task, which slows down the exploration speed.

\begin{figure*}[htbp]
      \vspace{2mm}
      \centering
      \includegraphics[width=0.67\linewidth, height=0.25\linewidth]{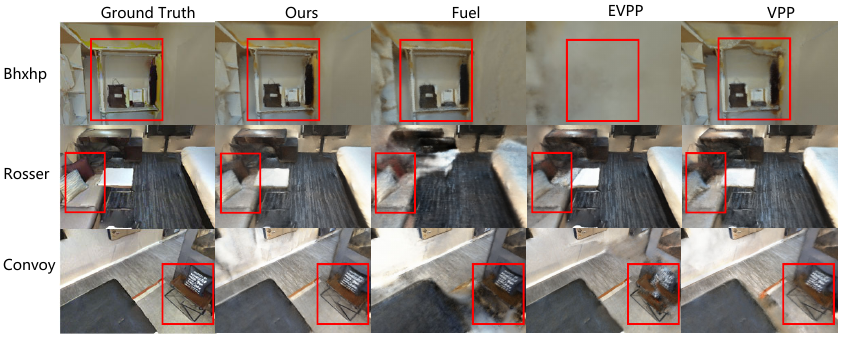}
    \includegraphics[trim=0cm 0cm 0cm 0.5cm, clip, width=0.67\linewidth,height=0.23\linewidth]{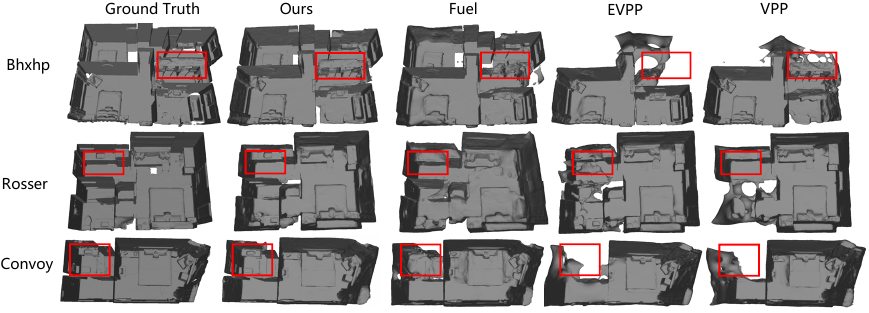}
      \caption{Comparison with different methods. Top: novel view synthesis; Bottom: reconstructed meshes.}
      \figlabel{geometry_compare}
      \vspace{-6mm}
\end{figure*}

\figref{geometry_compare} shows our method provides better reconstruction results in novel views and geometry. For more visual comparisons and results, we refer readers to the supplementary video. ~\figref{trajectory_compare} demonstrates that the trajectory of our method expands in scene $Bhxhp$ that of other methods.

\begin{figure}[htbp]
      \centering
      \includegraphics[width=0.9\linewidth, height=0.6\linewidth]{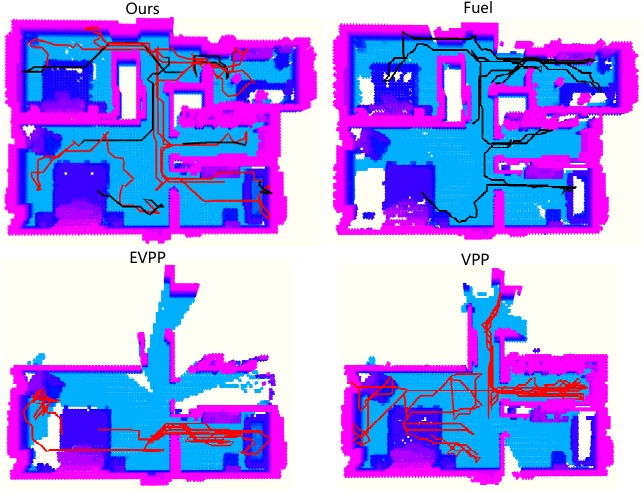}
      \caption{Comparison of trajectories with different methods.}
      \figlabel{trajectory_compare}
\vspace{-5mm}
\end{figure}


\subsection{Robot experiments in real scene}
We implemented our proposed method on an actual UAV equipped with Realsense Depth Camera D435i to perform room reconstruction, specifically targeting a room with dimensions of 8m × 2.5m × 3m. The pose of the camera is provided by Droid-SLAM~\cite{teed2021droid}. For this scene, the UAV takes about 5 minutes to explore and reconstruct the room. The exploration and reconstruction results will be presented in the supplementary video.

\seclabel{rers}

\section{Conclusion}

In this paper, we combine exploration and reconstruction tasks to ensure both global coverage and high-quality reconstruction. Subsequently, we introduce implicit surface uncertainty to accelerate view selection. Finally, we employ an adaptive mode switching method to improve planning efficiency without falling into local optimum. Comprehensive experiments demonstrate the superior performance of our method. 

In the future, we plan to study how viewpoint selections impact pose estimation in our planning process, while also extending our research to multi-agents.

{\small
\bibliographystyle{ieeefullname}
\bibliography{multi_task_planning}
}

\end{document}